\documentclass[letterpaper, 10 pt, conference]{ieeeconf}  

\IEEEoverridecommandlockouts                              

\usepackage{url}
\usepackage{comment}
\usepackage{amsmath,amssymb,amsfonts, mathtools}
\usepackage{graphicx}
\usepackage{bm}
\usepackage{epstopdf}
\usepackage{placeins}
\usepackage{bm}
\usepackage{cite}
\let\labelindent\relax
\usepackage[inline]{enumitem}
\usepackage{multicol}
\usepackage{booktabs}
\usepackage{multirow}
\usepackage{tikz}
\usepackage{caption}
\usepackage{subcaption}
\usepackage{tikz}
\usepackage{svg}
\usepackage[linesnumbered,ruled,vlined]{algorithm2e}
\SetKwInput{KwInput}{Input}    
\SetKwInput{KwOutput}{Output}  

\newcommand{\vect}[1]{\boldsymbol{#1}}
\newcommand\numberthis{\addtocounter{equation}{1}\tag{\theequation}}
\DeclareMathOperator*{\argmin}{arg\,min}

\title{\LARGE \bf Robot Learning Using Multi-Coordinate Elastic Maps}

\author{Brendan Hertel and Reza Azadeh
	\thanks{Authors are with the Persistent Autonomy and Robot Learning (PeARL) Lab, University of Massachusetts Lowell, Lowell, MA 01854, USA. Emails: \tt{brendan\_hertel@student.uml.edu, reza@cs.uml.edu}}
 }
 
\begin{document}

\maketitle
\thispagestyle{empty}
\pagestyle{empty}

\begin{abstract}
    To learn manipulation skills, robots need to understand the features of those skills. An easy way for robots to learn is through Learning from Demonstration (LfD), where the robot learns a skill from an expert demonstrator. While the main features of a skill might be captured in one differential coordinate (i.e., Cartesian), they could have meaning in other coordinates. For example, an important feature of a skill may be its shape or velocity profile, which are difficult to discover in Cartesian differential coordinate. In this work, we present a method which enables robots to learn skills from human demonstrations via encoding these skills into various differential coordinates, then determines the importance of each coordinate to reproduce the skill. We also introduce a modified form of Elastic Maps that includes multiple differential coordinates, combining statistical modeling of skills in these differential coordinate spaces. Elastic Maps, which are flexible and fast to compute, allow for the incorporation of several different types of constraints and the use of any number of demonstrations. Additionally, we propose methods for auto-tuning several parameters associated with the modified Elastic Map formulation. We validate our approach in several simulated experiments and a real-world writing task with a UR5e manipulator arm.
\end{abstract}

\section{Introduction}
\label{sec:intro}

Robots must be able to provide general use in a wide variety of applications, meaning that they must acquire new skills effortlessly. Pre-programming all necessary movements is infeasible, therefore robots should have methods to learn new skills on-the-fly. One of the most suitable methods for teaching robots new skills is Learning from Demonstration (LfD)~\cite{ravichandar2020recent}, where a human teacher demonstrates a skill which is then encoded by the robot. These skills can be performed using a variety of methods: a human user physically moving the robot as in kinesthetic teaching, an operator teleoperating the robot, or a robot watching a human perform the skill. This offers a low-effort and simply programmable method to capture demonstrations which can then be used by a robot learner.

For robots to best reproduce demonstrated skills, they should understand the \textit{meaning} behind them. For some skills, the information present in task space is important, for instance in a reaching skill, arriving at the endpoint is crucial. However, this may not be the case for all skills. When writing a letter, preserving the shape of the letter is more important than the physical position of the letter itself. Therefore, the robot should encode the desired skill in such a way that captures not just the positions of the skill, but also the importance of other features.

\begin{figure}[t]
    \centering
    \includegraphics[width=0.98\linewidth]{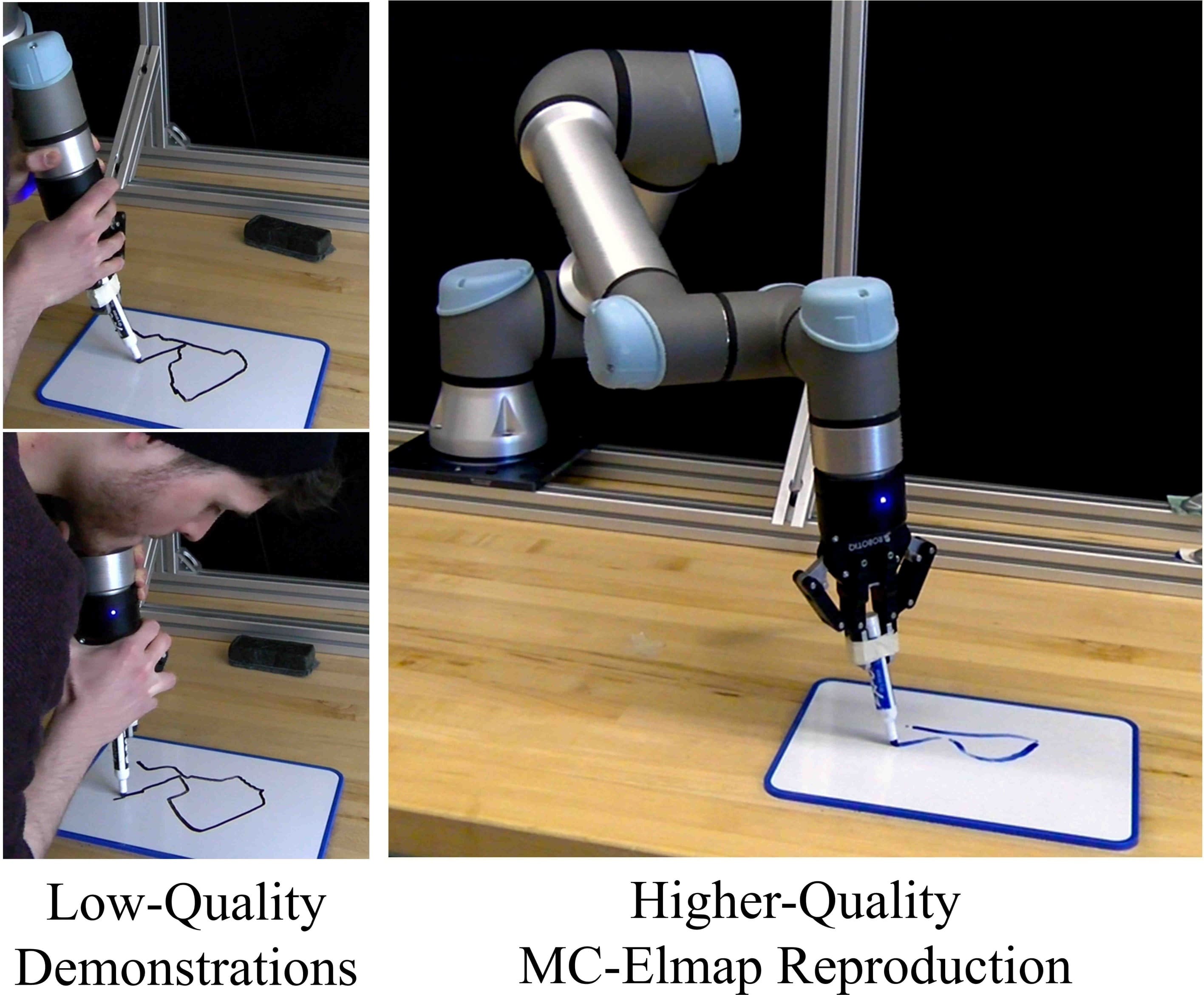}
    \caption{\small{Multi-Coordinate Elastic Maps (MC-Elmap) performed on a real-world writing skill. Multiple ``R'' shapes are drawn using a real-world UR5e manipulator arm, but the demonstrations are of low quality as it is difficult to apply enough pressure to write the shape while drawing smoothly. However, MC-Elmap is able to successfully and smoothly reproduce the skill, emphasizing features which were not present in the demonstrations.}}
    \label{fig:fig1}
\end{figure}

Recent advances in the field have introduced a variety of new Learning from Demonstration (LfD) representations.
One representation, Multi-Coordinate Cost Balancing (MCCB)~\cite{ravichandar2019skill}, encodes demonstrations into multiple differential coordinates, optimizes the importance of these coordinates, and then optimizes a reproduction according to the optimal importance weights. MCCB uses Gaussian Mixture Models and Gaussian Mixture Regression (GMM/GMR) to encode demonstrations. In this work, we use a statistical method which has recently seen many applications in LfD, known as Elastic Maps~\cite{hertel2022ElMap, hertel2023confidence, donald2024adaptive} to encode demonstrations. Our proposed method, which we denote Multi-Coordinate Elastic Maps (MC-Elmap), provides several advantages over MCCB. First, Elastic Maps have a cost term for smoothness, leading to smoother reproductions in the event of noisy demonstrations, as seen in Fig.~\ref{fig:fig1}. Second, Elastic Maps are extremely flexible, and can be used with any number of demonstrations. The contributions of this work are as follows:
\begin{enumerate*}[label=(\roman*)]
    \item a novel Elastic Maps formulation which includes the ability to incorporate features from multiple differential coordinates,
    \item methods for auto-tuning hyperparameters, allowing for ease of use and application, and
    \item comparisons with several baselines.
\end{enumerate*}
We validate the use of MC-Elmap in 28 simulated and one real-world experiment.

\section{Related Work}
\label{sec:RW}

There are many potential approaches to Learning from Demonstration (LfD)~\cite{li2024learning}, all of which can roughly be grouped into four main categories: dynamical systems~\cite{pastorDMP2009}, probabilistic~\cite{Paraschos2013ProMP}, geometric~\cite{Ahmadzadeh2018TLGC}, and statistical~\cite{Calinon2007GMM} methods. Each category presents its own unique advantages, and many methods combine these groupings for best results~\cite{figueroa2018physically}. Dynamical systems attempt to create a stable function over the workspace, resulting in the ability to start from anywhere in said workspace~\cite{Khansari-Zadeh2011LASA}. Probabilistic models find a ``likely execution'' for a skill and can be highly flexible~\cite{huang2019kernelized}. Geometric models attempt to capture the geometric features of a skill, which is highly useful for geometrically defined movements, or skills where certain features remain invariant across transformations~\cite{Nierhoff2016LTE}. Finally, statistical methods find statistical representations of skills, which can often be used as encodings for combining multiple skills or features~\cite{hertel2021TLFSD}. In this work, we combine statistical modeling of skills in multiple geometrically-defined differential coordinate spaces, allowing us to capture geometric features across many different demonstrations and skills. 

Some LfD methods specifically encode skills in different spaces. One of these methods is Laplacian Trajectory Editing (LTE)~\cite{Nierhoff2016LTE}. LTE transforms demonstrations into Laplacian coordinate space, adds task constraints, then transforms them back to Cartesian coordinate space. This results in a reproduction which follows the shape of the demonstrations while adhering to any number of initial, final, or via-point constraints. One drawback of this method is that it only uses Laplacian space, and does not consider other coordinate spaces which may capture features of the skill. Additionally, LTE only works with single demonstrations, offering no ability to encode skill variances measured from multiple demonstrations. Building upon LTE, Multi-Coordinate Cost Balancing (MCCB)~\cite{ravichandar2019skill} remedies some of these drawbacks.  MCCB uses Gaussian Mixture Models and Gaussian Mixture Regression (GMM/GMR)~\cite{Calinon2007GMM} to model robot skills (although any similar encoding could be used). Additionally, MCCB uses Cartesian, Tangent, and Laplacian differential coordinates, utilizing a meta-optimization to balance the importance of each coordinate according to the skill. Trajectory Learning via Failed and Successful Demonstrations (TLFSD)~\cite{hertel2021TLFSD} applies the ideas of MCCB to learn the features specific to success or failure of a skill in multiple differential coordinate spaces. Alternatively, Elastic-Laplacian Trajectory Editing (ELTE)~\cite{donald2024adaptive} applies the ideas from LTE in an online adaptation scenario, where reproductions must be updated during execution while maintaining the shape of the demonstrations. Our novel approach, MC-Elmap, addresses the main limitations of previous methods by integrating their most essential features. MC-Elmap can incorporate either single or multiple demonstrations, satisfy any number of initial, final, or via-point constraints, automatically determine importance of different coordinate spaces, and be updated for online execution.

Skill learning methods that rely on  Reinforcement Learning (RL)~\cite{singh2022reinforcement} employ an agent which explores different rollouts through the environment, receiving a higher reward for better executions. With RL, the skill must be defined according to specific features, which can include geometric features captured in different coordinate spaces. However, these features may be difficult to define manually, especially with regards to their relative importance in the skill. To this end, Inverse Reinforcement Learning (IRL)~\cite{abbeel2004apprenticeship} offers a solution. In IRL, the features of the skill are predetermined, but their relative importance is unknown. Given some expert examples of the skill, an importance can be found which optimizes the performance of these examples. Similarly to IRL, MC-Elmap uses expert examples to determine the relative importance of features captured in various differential coordinate spaces of a demonstrated skill.

\section{Methodology}
\label{sec:method}

\subsection{Background on Elastic Maps}

First, we briefly explain Elastic Maps~\cite{gorban2005elastic, hertel2022ElMap}. Elastic Maps model trajectories as a series of nodes connected via springs. Nodes are connected to the demonstration data through springs, as well as adjacent nodes. Adjacent node connections make edges, and pairs of edges create ribs. Minimizing the spring energies associated with the Elastic Map results in an optimized map, which was shown to be well-suited for replicating robot trajectories~\cite{hertel2022ElMap}.

Typically, the energies for the Elastic Maps are as follows: 
\begin{enumerate*}[label=(\roman*)]
    \item the approximation energy $u_\mathcal{X}$, which promotes a good fit to the demonstration data
    \item the stretching energy $u_E$, which promotes equally-spaced nodes, and
    \item the bending energy $u_R$, which promotes straightness of nodes (the stretching and bending energies together promote smoothness in the result).
\end{enumerate*}

\subsection{Multi-Coordinate Elastic Maps}

In \cite{donald2024adaptive}, the approximation energy, $u_\mathcal{X}$, was modified such that it approximated the shape of the demonstration rather than converging to the demonstration itself. We take this idea one step further and propose a novel formulation of Elastic Maps inspired by MCCB~\cite{ravichandar2019skill}. In MCCB, the Cartesian (zero-th order), Tangent (first order), and Laplacian (second order) differential coordinate frames of the demonstrations are balanced according to several hyperparameters. We first are given $N \geq 1$ demonstrations as a set $\mathcal{D} = \{ \vect{D}_1, \vect{D}_2, \dots, \vect{D}_N \}$, where a single time-aligned demonstration is defined as $\vect{D}_i = [ \vect{x}_i^1, \vect{x}_i^2, \dots, \vect{x}_i^T ]^\top \in \mathbb{R}^{T \times d}$, with a single data point $\vect{x}_i^j$ is a $d$-dimensional point in space. To convert the demonstrations into different differential coordinates, we use coordinate transformation matrices as 
\begin{align}
    \vect{D}_{\mathcal{T}, i} &= \vect{\mathrm{T}}\vect{D}_i, \label{eq:DT}\\  
    \vect{D}_{\mathcal{L}, i} &= \vect{\mathrm{L}}\vect{D}_i, \label{eq:DL}
\end{align}
where $\vect{\mathrm{T}}$ is the Graph Tangent matrix, and $\vect{\mathrm{L}}$ is the Graph Laplacian matrix, which take the following forms, respectively:

\begin{equation}
\vect{\mathrm{T}} = 
\begin{bmatrix}
 0 & 0 &  0 & \cdots & 0 \\
 -1 & 1 &  0 & \cdots & 0 \\
0 &  -1 & 1 & \cdots & 0 \\
\vdots & \ddots &\ddots & \ddots & \vdots \\
0 & \cdots & 0 & -1 & 1   \\
\end{bmatrix}, \nonumber
\end{equation}
\begin{equation}
\vect{\mathrm{L}} = \frac{1}{2}
\begin{bmatrix}
-2 &  2 & 0 & 0 & \cdots & 0 \\
1 & -2 &  1 & 0 & \cdots & 0 \\
0 & 1 & -2 & 1 & \cdots & 0 \\
\vdots & \ddots & \ddots & \ddots & \ddots & \vdots\\
0 & \cdots & 0 & 1 & -2 & 1   \\
0 & \cdots & 0 & 0 & 2 & -2    \\
\end{bmatrix}. \nonumber
\end{equation}
where both matrices are square~\cite{ravichandar2019skill}. Elastic Maps fit all data at once (i.e., batch learning), therefore we concatenate the demonstrations as $\vect{g} = [ \vect{x}_1^1, \vect{x}_2^1, \dots, \vect{x}_N^T ]$ in Cartesian space, and do the same for Tangent and Laplacian spaces, resulting in $\vect{g}_\mathcal{T}$ and $\vect{g}_\mathcal{L}$, respectively. 

Typically, Elastic Maps use three energy penalty costs in order to find an optimized map: the Cartesian approximation energy, the stretching energy, and the bending energy~\cite{hertel2022ElMap}. These energies are formulated as follows:
\begin{align}
    u_\mathcal{X} &= w_\mathcal{X} || \vect{W}\vect{y} - \vect{K}\vect{g} ||_2^2, \label{eq:U_X1}\\
    u_E &= \lambda || \vect{E}\vect{y} ||_2^2,  \label{eq:U_E1}\\
    u_R &= \mu || \vect{R}\vect{y} ||_2^2 \label{eq:U_R1}
\end{align}
where $u_\mathcal{X}$ is the Cartesian approximation energy, defined as the difference in Cartesian space between the demonstration data $\vect{g}$ and the reproduction $\vect{y} \in \mathbb{R}^{M \times d}$. Since $|\vect{g}|$ (where $|\cdot|$ represents cardinality) may be greater than $|\vect{y}|$, a clustering is needed to determine which data point a node in $\vect{y}$ represents. A data point  $\vect{g}_j$ is clustered to node $\vect{y}_i$ if there exists no other node $\vect{y}_k$ closer in Euclidean space. This clustering is used to find the clustering matrix $\vect{K}$ and weighting matrix $\vect{W}$. The clustering matrix $\vect{K}$ is defined such that if data point $\vect{g}_j$ is clustered to node $\vect{y}_i$ (i.e., $\vect{g}_j \in \kappa_i$, where $\kappa_i$ is the cluster corresponding to $\vect{y}_i$), then $\vect{K}_{ij} = 1$, otherwise $\vect{K}_{ij} = 0$. $\vect{W}$ is a diagonal weighting matrix such that $\vect{W}_{ii} = \sum_j \vect{K}_{ij}$. The weighting constant $w_\mathcal{X}$ defines the importance of the Cartesian approximation energy, and $||\cdot||_n$ is the $L^n$-norm. Additionally, the stretching energy $u_E$ and bending energy $u_R$ find the energies from the edges and ribs, respectively. The edge matrix $\vect{E}$ and rib matrix $\vect{R}$ are defined as:
\begin{equation}
\vect{E} = 
\begin{bmatrix}
 -1 & 1 &  0 & \cdots & 0 \\
0 &  -1 & 1 & \cdots & 0 \\
\vdots & \ddots &\ddots & \ddots & \vdots \\
0 & \cdots & 0 & -1 & 1   \\
\end{bmatrix}, \nonumber 
\end{equation}
\begin{equation}
\vect{R} = 
\begin{bmatrix}
1 & -2 &  1 & 0 & \cdots & 0 \\
0 & 1 & -2 & 1 & \cdots & 0 \\
\vdots & \ddots & \ddots & \ddots & \ddots & \vdots\\
0 & \cdots & 0 & 1 & -2 & 1   \\
\end{bmatrix}. \nonumber
\end{equation}
and the parameters $\lambda$ and $\mu$ are weighting constants for the stretching and bending energies, respectively. 

In this novel Elastic Map formulation, we incorporate approximation energy in the Tangent and Laplacian coordinate spaces as well. The energy terms associated with these are as follows:
\begin{align}
    u_\mathcal{T} &= w_\mathcal{T}  || \vect{W}\vect{y}_\mathcal{T} - \vect{K}\vect{g}_\mathcal{T} ||_2^2, \label{eq:U_T1}\\
    u_\mathcal{L} &= w_\mathcal{L}  || \vect{W}\vect{y}_\mathcal{L} - \vect{K}\vect{g}_\mathcal{L} ||_2^2, \label{eq:U_L1}
\end{align}
where $u_\mathcal{T}$ is the Tangent approximation energy, defined as the difference in Tangent space between the demonstration data and the reproduction, weighted according to the parameter $w_\mathcal{T}$, and $u_\mathcal{L}$ is the Laplacian approximation energy, defined as the difference in Laplacian space between the demonstration data and the reproduction, weighted according to the parameter $w_\mathcal{L}$. The reproduction in Tangent space, $\vect{y}_\mathcal{T}$, and Laplacian space, $\vect{y}_\mathcal{L}$, can be found using the same transformation method as the demonstrations in \eqref{eq:DT} and \eqref{eq:DL}, respectively. 

By minimizing the energy terms in \eqref{eq:U_X1}-\eqref{eq:U_L1} as
\begin{align*}
    &\underset{y}{\text{minimize }} f_0(y) = \sum_{i \in \{\mathcal{X}, \mathcal{T}, \mathcal{L}, E, R \} } u_i \numberthis\label{eq:opt} \\
    &\text{subject to} \ f_i(y) = ||\vect{y}_j - \vect{z}||_1 \leq 0
\end{align*}
we find an optimal Multi-Coordinate Elastic Map reproduction. This optimization can include any number of initial, final, or via-point constraints in $f_i(y)$, which are defined such that a point along the reproduction $\vect{y}_j$ is constrained to some point in space $\vect{z}$. Additionally, \eqref{eq:opt} is convex, meaning it can be solved quickly and efficiently~\cite{hertel2023confidence}. Another benefit of this formulation is that this optimization can also be modified for online trajectory adaptation~\cite{donald2024adaptive}. Since this optimization incorporates a clustering process, finding an optimal reproduction may change the clustering. Therefore, we can use an Expectation-Maximization (EM) algorithm to iteratively update the clustering (E-step) and solve for an optimal reproduction (M-step) until convergence. Convergence occurs when the clustering does not change between iterations, resulting in a local optimum solution. Psuedocode for this algorithm can be seen in Algorithm 1. 

\begin{algorithm}[t]
\DontPrintSemicolon
\label{MCE_alg}
  \KwInput{$M$, Initial Nodes $\vect{y}$, Data $\mathcal{D}$, Parameters $\lambda_0, \mu_0$}
  \KwOutput{Optimal Nodes $\vect{y}$*}

   Initialize $\vect{\mathrm{T}}, \vect{\mathrm{L}}, \vect{E}, \vect{R}$\; 
   $\vect{g} \longleftarrow \text{CONCATENATE}(\mathcal{D})$\;
   $\vect{g}_\mathcal{T} \longleftarrow \text{CONCATENATE}(\mathcal{D}_\mathcal{T})$\;
   $\vect{g}_\mathcal{L} \longleftarrow \text{CONCATENATE}(\mathcal{D}_\mathcal{L})$\;
   \While{not converged}
   {
        \tcp{Expectation}
        $\kappa_i$ = [] for $i = 1...M$\;
        
        \For{$\vect{g}_i \in \vect{g}$}
        {
            $k = \argmin_{j=1...M} ||\vect{g}_i - \vect{y}_j||$\;
            $\kappa_k \longleftarrow i$\;
        }
        \tcp{Maximization}
        $\vect{K}$ = []\;
        $\vect{W}$ = []\;
        \For{$i=1...M$}
        {
            $\vect{K}_{ij} = 1$ for $j \in \kappa_i$\;
            $\vect{W}_{ii} = \sum_{j} \vect{K}_{ij}$\;
        }
        \tcp{Hyperparameter Tuning}
        $[\beta_\mathcal{X}, \beta_\mathcal{T}, \beta_\mathcal{L}]$ = CALC-SCALING($\vect{y}, \mathcal{D}$) \ \ \tcp{eq. \eqref{eq:betas}}
        $[\alpha_\mathcal{X}, \alpha_\mathcal{T}, \alpha_\mathcal{L}]$ = minimize $\sum_{j=1}^N J_\mathcal{X}(\vect{D}_j) $\;
        $[w_\mathcal{X}, w_\mathcal{T},w_\mathcal{L}]$ = [$\frac{\alpha_\mathcal{X}}{\beta_\mathcal{X}}, \frac{\alpha_\mathcal{T}}{\beta_\mathcal{T}}, \frac{\alpha_\mathcal{L}}{\beta_\mathcal{L}}$]\;
        [$\lambda, \mu$] = CALC-SMOOTHING($\vect{y}, \mathcal{D}, \lambda_0, \mu_0$) \tcp{eqs. \eqref{eq:calc-lambda}, \eqref{eq:calc-mu}}
        \tcp{Optimization}
        $\vect{y}$* = minimize $\sum_{i \in \{\mathcal{X}, \mathcal{T}, \mathcal{L}, E, R \} } u_i$
        
   }

\caption{MC-Elmap Expectation-Maximization Algorithm}
\end{algorithm}

\subsection{Hyperparameter Optimization}

Different values of the weighting parameters lead to very different results of \eqref{eq:opt}. Therefore, an optimal set of parameters must be found for each reproduction. To optimize the parameters we perform a meta-optimization on \eqref{eq:opt}. First, the weight values of the approximation terms are estimated as 
\begin{align}
    \hat{w}_\mathcal{X} = \frac{\alpha_\mathcal{X}}{\beta_\mathcal{X}}, \
    \hat{w}_\mathcal{T} = \frac{\alpha_\mathcal{T}}{\beta_\mathcal{T}}, \
    \hat{w}_\mathcal{L} = \frac{\alpha_\mathcal{L}}{\beta_\mathcal{L}}, \nonumber
\end{align}
where $\alpha_i$ are positive weight values to determine the cost of the corresponding differential coordinate satisfying $\sum_i \alpha_i = 1$; and $\beta_i$ are scaling factors for the corresponding differential coordinates satisfying $\sum_i \beta_i = 1$. The scaling factors are determined by
\begin{equation}
    \beta_i = \frac{\sum_{j=1}^N J_i (\vect{D}_j)}{\sum_k \sum_{j=1}^N J_k (\vect{D}_j)}, \ k=\{\mathcal{X}, \mathcal{T}, \mathcal{L}\} \label{eq:betas}
\end{equation}
where $J$ is the cost associated with the unweighted portion of each approximation energy term, i.e., $J_\mathcal{X} = || \vect{W}_i\vect{y} - \vect{K}_i\vect{D}_i ||_2^2$. Note that we use a clustering and weighting matrix for each individual demonstration, denoted $\vect{K}_i$ and $\vect{W}_i$ respectively, which can be found using the same process as the full clustering and weighting matrices. The cost of each differential coordinate space can be found by minimizing the reproduction error as 
\begin{equation}
    \{\alpha_\mathcal{X}, \alpha_\mathcal{T}, \alpha_\mathcal{L}\} = \underset{\{\alpha_\mathcal{X}, \alpha_\mathcal{T}, \alpha_\mathcal{L}\}}{\text{minimize }} \sum_{j=1}^N J_\mathcal{X}(\vect{D}_j) \label{eq:alphas}
\end{equation}
where $y$ is found by solving \eqref{eq:opt}.

Next, we propose a method for tuning the stretching and bending parameters $\lambda$ and $\mu$. We base this tuning on the assumption that the approximation energies should be about equal to the stretching and bending energies, that is $u_{\mathcal{X}} + u_{\mathcal{T}} + u_{\mathcal{L}} \approx u_E \approx u_R$. Therefore, we use an initial guess for $\vect{y}$ and calculate the approximation energies. Then we calculate the bending and stretching energies, unmodified by the constants. To find the parameter $\lambda$ we apply the following
\begin{equation}
    \lambda = \lambda_0 \frac{u_{\mathcal{X}} + u_{\mathcal{T}} + u_{\mathcal{L}}}{||\vect{E}\vect{y}||_2^2} \label{eq:calc-lambda}
\end{equation}
where $\lambda_0 > 1$ is a small constant. The process is the same for finding $\mu$, but replacing $E$ with $R$, as
\begin{equation}
    \mu = \mu_0 \frac{u_{\mathcal{X}} + u_{\mathcal{T}} + u_{\mathcal{L}}}{||\vect{R}\vect{y}||_2^2} \label{eq:calc-mu}
\end{equation}
where $\mu_0 > 1$ is another small constant. The pseudo-algorithm for the Expectation-Maximization algorithm of Multi-Coordinate Elastic Maps (MC-Elmap) along with these parameter tunings (lines 15-18) can be seen in Algorithm~1.\footnote{Available at: \url{https://github.com/brenhertel/MC-Elmap}}

\begin{figure}[t]
    \centering
    \includegraphics[width=0.98\linewidth]{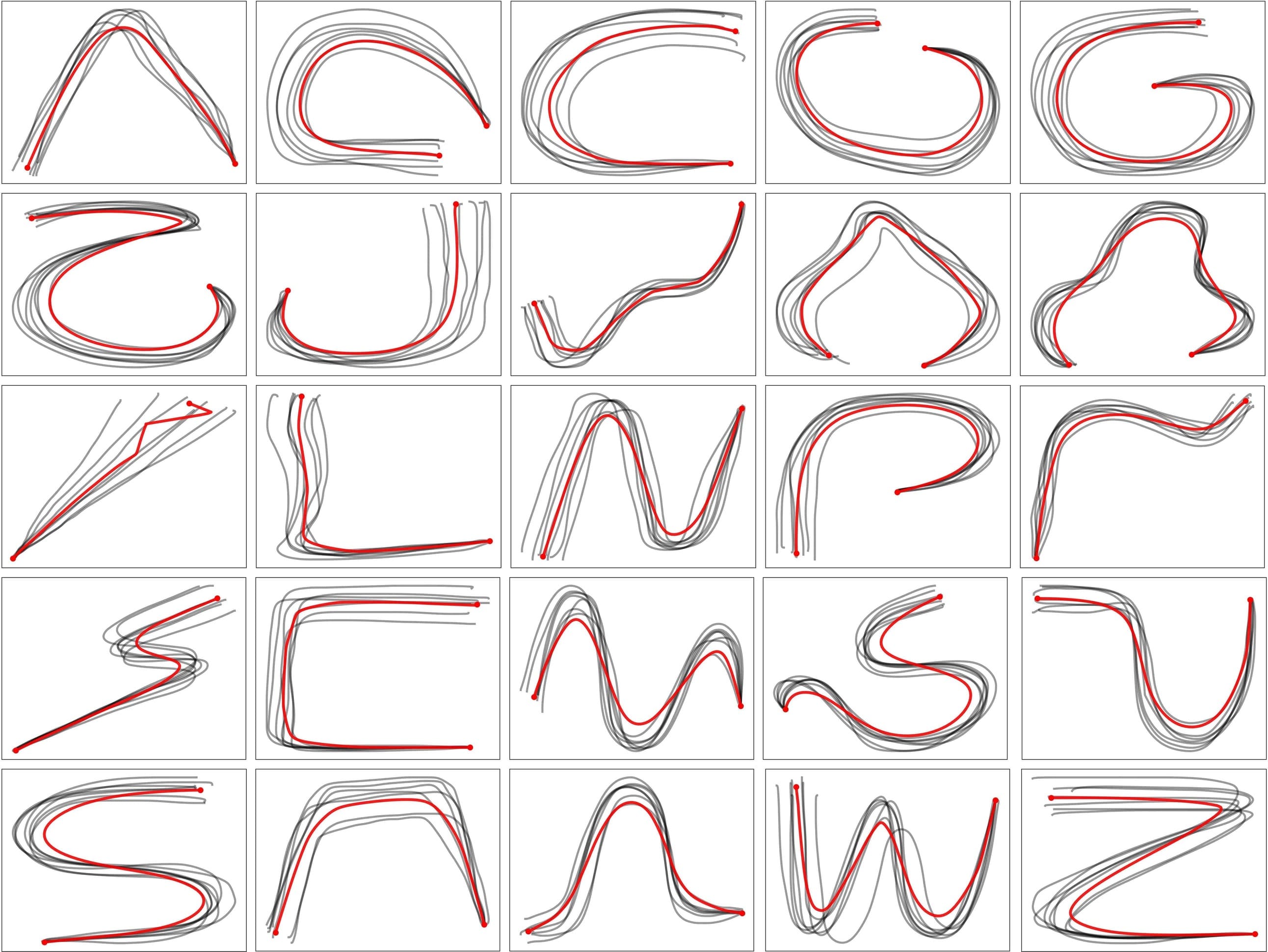}
    \caption{\small{Results of performing MC-Elmap (red) on the handwriting shapes LASA dataset~\cite{Khansari-Zadeh2011LASA} (demonstrations shown in gray).}}
    \label{fig:lasa}
\end{figure}

\begin{figure*}[bt]
    \centering
    \includegraphics[width=0.98\linewidth]{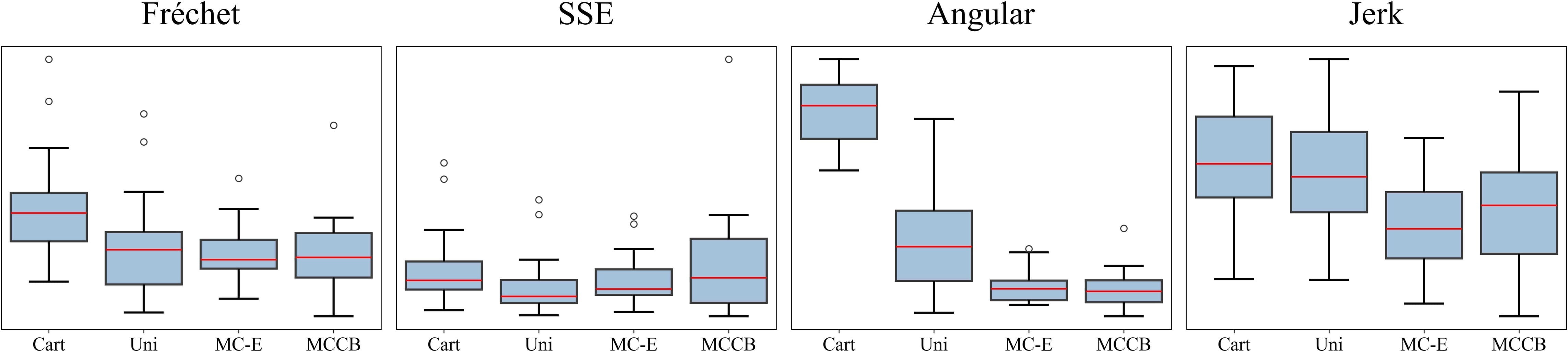}
    \caption{\small{Boxplots comparing the results of measuring the Fr\'echet distance, Sum of Squared Errors (SSE), Angular Similarity, and Jerk for different LfD representations on the LASA Dataset (see Fig.~\ref{fig:lasa}). MC-Elmap (denoted MC-E in the plots) performs well for all metrics. Boxplot whiskers show 1.5 interquartile range (IQR), with the median shown in red.}}
    \label{fig:boxplots}
\end{figure*}

\begin{figure}[ht]
    \centering
    \includegraphics[width=0.981\linewidth]{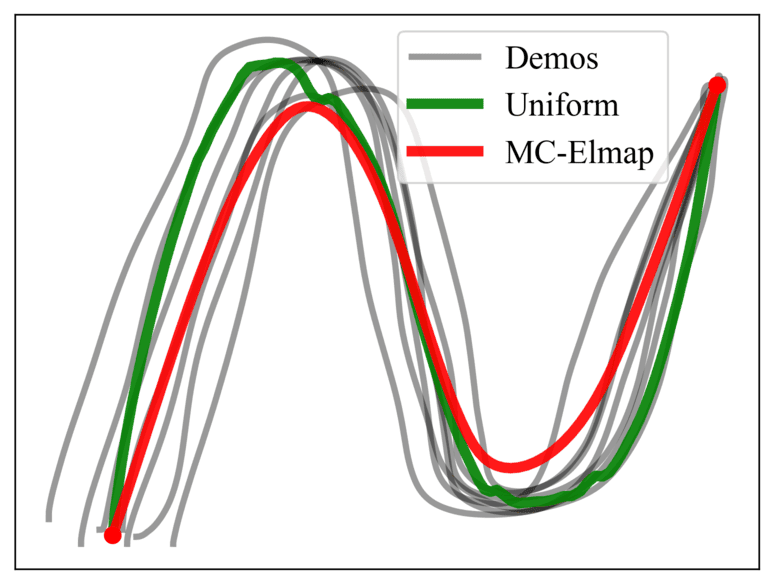}
    \caption{\small{A comparison between uniform weighting (shown in green) and automatic weighting (shown in red) for MC-Elmap on the ``NShape'' of the LASA dataset (demonstrations shown in gray). Uniform weighting places a higher importance on Cartesian similarity, leading to some jaggedness the reproduction in order to increase spatial similarity in the reproduction, whereas the automatic weighting places a higher importance on Laplacian similarity, maintaining the shape of the handwriting skill.}}
    \label{fig:nshape}
\end{figure}

\begin{figure*}[bt]
    \centering
    \includegraphics[width=0.98\linewidth]{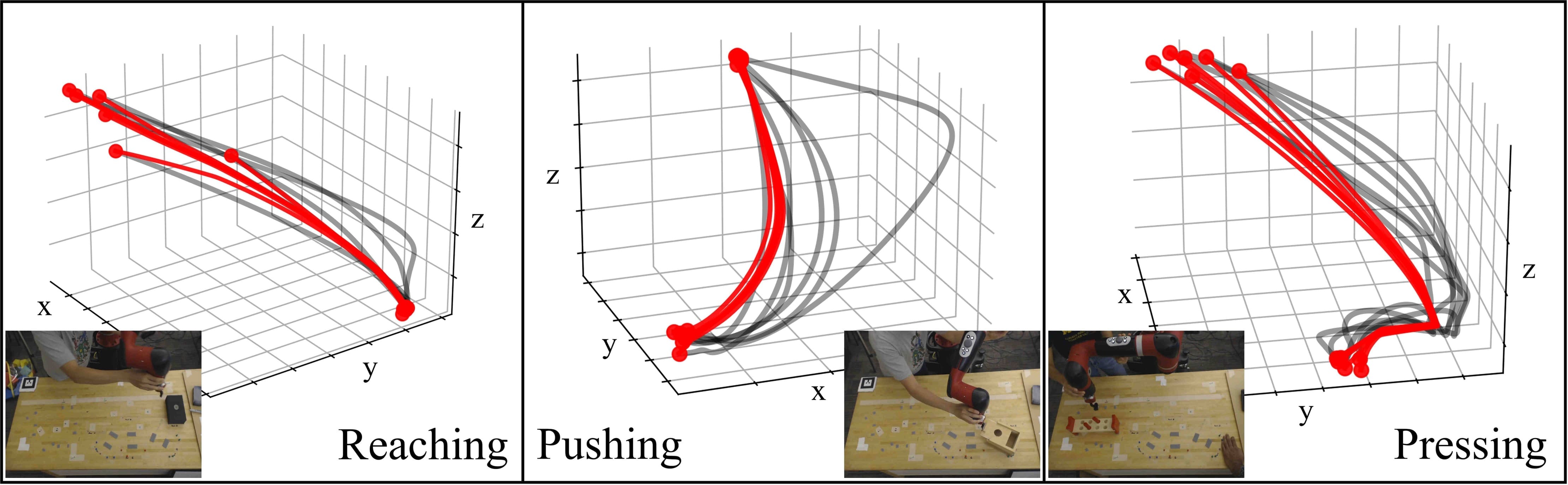}
    \caption{\small{Results of performing MC-Elmap (red) on the RAIL dataset~\cite{rana2020benchmark} (demonstrations in gray). A sample real-world demonstration of each skill is shown in the bottom corner for each skill. MC-Elmap is able to successfully reproduce all skills from multiple different starting points, and even is able to reproduce features without constraints, such as the first press in the pressing task.}}
    \label{fig:rail}
\end{figure*}

\section{Experiments}
\label{sec:exp}

We perform several experiments to evaluate the performance of Multi-Coordinate Elastic Maps (MC-Elmap). We first use MC-Elmap on each shape in the LASA dataset~\cite{Khansari-Zadeh2011LASA}, a 2D handwriting dataset consisting of 25 shapes (JShape2 has been omitted), shown in Fig.~\ref{fig:lasa}. MC-Elmap finds smooth reproductions which maintain the shape of all letters, an important feature in the handwriting task. Additionally, we measure the Fr\'echet distance, Sum of Squared Errors (SSE), Angular Similarity, and overall jerk of the reproductions (lower is better for all metrics). The Fr\'echet distance measures spatial but not temporal similarity, SSE measures spatial and temporal similarity, Angular Similarity measures the similarity in shape between two curves, and jerk measures the smoothness of the reproduction. The boxplots of these metrics are shown in Fig.~\ref{fig:boxplots}, along with comparisons to the following methods:
\begin{itemize}[leftmargin=*]
    \item Cartesian-only Elmap (Cart)~\cite{hertel2022ElMap} ($w_\mathcal{X} = 1, w_\mathcal{T} =  w_\mathcal{L} = 0$),
    \item Uniform MC-Elmap (Uni) ($w_\mathcal{X} =  w_\mathcal{T} = w_\mathcal{L} = 0.33$), and
    \item MCCB~\cite{ravichandar2019skill}.
\end{itemize}
As can be seen from these comparisons (with mean values across all shapes, normalized by the worst-performing result, listed in Table~\ref{tab:lasa-table}), MC-Elmap finds consistently excellent results. As shown in Fig.~\ref{fig:boxplots}, for each metric, MC-Elmap outperforms other methods. MC-Elmap and MCCB both perform the best on average for the Fr\'echet distance metric, indicating a strong ability to maintain spatial similarity to demonstrations. This makes sense, as both methods use an optimization which rewards spatial similarity. For SSE, Uniform MC-Elmap performs the best, with automatic weighting MC-Elmap performing the second-best. The reduced performance of auto-weighting could be due to the fact that the auto-weighting favors other factors besides spatial and temporal similarity, such as smoothness. For Angular Similarity, MCCB performs the best, with MC-Elmap showing very similar performance. Finally, MC-Elmap performs the best on jerk, indicating a strong preference for smoothness, which is an important feature for robot trajectories. A comparison between the results of uniform weighting and automatic weighting can be seen in Fig.~\ref{fig:nshape}. In this handwriting task, the shape of the demos is most important to the task, and intuitively the Laplacian approximation energy should have more importance. We find that automatic weighting does weight Laplacian approximation more importantly ($\hat{w}_\mathcal{X} = 0.01, \hat{w}_\mathcal{T} = 0.07, \hat{w}_\mathcal{L} = \bf{0.92}$), leading to better results than uniform weighting. A high weight for the Cartesian approximation term results in the reproduction introducing jaggedness (as seen around the bends of the ``NShape'') which is not seen with automatic weighting.

\begin{table}[t]
\centering
\caption{\small{Quantitative metric results from evaluating several methods across all shapes in the LASA dataset (see Fig.~\ref{fig:lasa}).}}
\label{tab:lasa-table}
\begin{tabular}{@{}crrrr@{}}
\toprule
\multicolumn{1}{l}{} & \multicolumn{1}{c}{Cartesian} & \multicolumn{1}{c}{Uniform} & \multicolumn{1}{c}{MC-Elmap} & \multicolumn{1}{c}{MCCB} \\ \midrule
Fr\'echet  & 0.50 & 0.37 & \textbf{0.35} & \textbf{0.35}  \\
SSE  & 0.20 & \textbf{0.13} & 0.16 & 0.22  \\
Angular  & 0.84 & 0.36 & 0.22 & \textbf{0.20}  \\
Jerk  & 0.70 & 0.73 & \textbf{0.56} & 0.61 \\ \bottomrule
\end{tabular}
\end{table}

Additionally, we use MC-Elmap on several 3D robot skills, namely the pressing, pushing, and reaching skills provided by the RAIL dataset~\cite{rana2020benchmark}. In this dataset, users perform these skills from several starting positions. We highlight the generalizability of this method, showcasing reproductions from several starting positions as seen in Fig.~\ref{fig:rail}. MC-Elmap is trained on all demonstrations shown, and a reproduction is generated using the initial and final constraints of each individual demonstration. MC-Elmap is able to successfully reproduce all skills from multiple starting points, following trajectories and reproducing the shape of demonstrations while maintaining smoothness. Additionally, MC-Elmap is able to discover important features, such as the first button press in the ``pressing'' skill, without the need for explicit constraints. In all reproductions, only the start and end point are constrained, although via-points could be used as constraints as well.

\begin{figure}[t]
    \centering
    \includegraphics[width=0.98\linewidth]{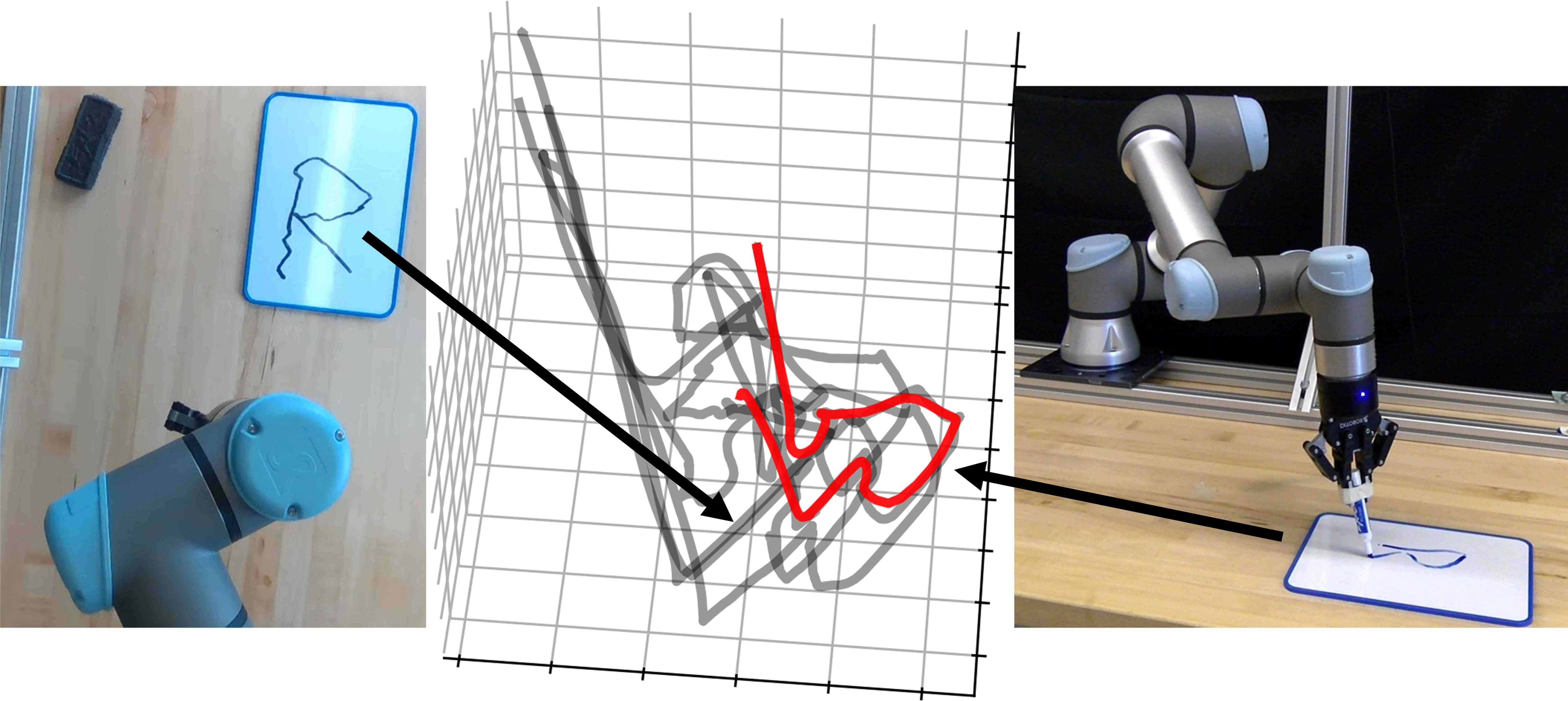}
    \caption{\small{Results of performing MC-Elmap (red) on a real-world writing task using a UR5e manipulator arm. Demonstrations are shown in gray. Demonstrations complete the task but with undesirable levels of jaggedness. MC-Elmap is able to find a smooth reproduction, resulting in a reproduction which more accurately resembles the intent of the task.}}
    \label{fig:writing3d}
\end{figure}

Finally, we perform MC-Elmap on a real-world handwriting task using a Universal Robots UR5e manipulator arm (as seen in Fig.~\ref{fig:fig1}). This arm was guided using kinesthetic teaching to draw an ``R'' shape on a whiteboard placed in the robot workspace. Demonstrations were of low quality, resulting in jagged letters, as can be seen from the demonstrations in Fig.~\ref{fig:writing3d}. This is due to the difficulty of applying enough pressure to write the letter on the whiteboard in combination with the coarse control kinesthetic teaching offers. However, MC-Elmap is able to smooth out the jaggedness, as well as discover the importance of shape for this handwriting task, and successfully reproduce the ``R'' shape, resulting in a higher-quality reproduction.\footnote{Accompanying video: \url{https://youtu.be/KU-ldkTa9UE}} This shows MC-Elmap is able to introduce desirable features into reproductions, such as smoothness, which are not necessarily present in demonstrations. Reproductions are computed \textit{a priori} then followed by a low-level controller, with rotations computed using Slerp~\cite{Shoemake1985slerp}.

\section{Conclusions \& Future Work}
\label{sec:cfw}

In this work we have presented Multi-Coordinate Elastic Maps (MC-Elmap), a method which balances the importance of several differential coordinates, then combines them to find an optimal reproduction. These reproductions can be auto-tuned to incorporate coordinates with more importance without the need for user input. Additionally, the Elastic Map modeling results in a smooth and evenly-spaced reproduction, which is well-suited for robot manipulation. We have presented a novel Multi-Coordinate formulation of Elastic Maps and methods to autotune several parameters associated with this formulation. We present this approach with several 2D and 3D datasets, as well as a real-world 3D handwriting task.

There are several avenues for future work. Currently, we use the Cartesian, Tangent, and Laplacian differential coordinates, but this could easily be expanded to include more frames. Alternatively some way of automatically determining the best frames to capture the features of the demonstration could be employed, similar to kernel-based methods. Furthermore, One of the features of Elastic Maps is that it can include variable weighting for each spring used in its energy calculation~\cite{gorban2005elastic}, but here we assume each spring has the same spring constant. Methods could be found to automatically and variably weight the springs, especially in areas of more importance within a skill.

\section*{Acknowledgments}

This research is supported in part by the National Science Foundation (FRR-2237463).

\typeout{}
\bibliographystyle{IEEEtran}
\bibliography{references}

\end{document}